\def\year{2019}\relax
\documentclass[letterpaper]{article} 

\usepackage{aaai19_aa}  
\usepackage{times}  
\usepackage{helvet}  
\usepackage{courier}  
\usepackage{url}  
\usepackage{graphicx}  

\usepackage{bm}
\usepackage{amsmath}
\usepackage{amssymb}
\usepackage{amsfonts}
\def\vec#1{\mathbf{#1}}

\usepackage{comment}	
\begin{document}
%
\title{Supervised Anomaly Detection based on Deep Autoregressive Density Estimators}
\author{Tomoharu Iwata \\ NTT Communication Science Laboratories \And 
Yuki Yamanaka \\ NTT Secure Platform Laboratories}

\maketitle
 \begin{abstract}
We propose a supervised anomaly detection method based on neural density estimators, where the negative log likelihood is used for the anomaly score. Density estimators have been widely used for unsupervised anomaly detection. By the recent advance of deep learning, the density estimation performance has been greatly improved. However, the neural density estimators cannot exploit anomaly label information, which would be valuable for improving the anomaly detection performance. The proposed method effectively utilizes the anomaly label information by training the neural density estimator so that the likelihood of normal instances is maximized and the likelihood of anomalous instances is lower than that of the normal instances. We employ an autoregressive model for the neural density estimator, which enables us to calculate the likelihood exactly. With the experiments using 16 datasets, we demonstrate that the proposed method improves the anomaly detection performance with a few labeled anomalous instances, and achieves better performance than existing unsupervised and supervised anomaly detection methods.
 \end{abstract}
 
\section{Introduction}

Anomaly detection is an important task in artificial intelligence,
which is a task to find anomalous instances in a dataset.
The anomaly detection has been used
in a wide variety of applications~\cite{chandola2009anomaly,patcha2007overview,hodge2004survey},
such as
network intrusion detection for cyber-security~\cite{dokas2002data,yamanishi2004line},
fraud detection for credit cards~\cite{aleskerov1997cardwatch},
defect detection of industrial machines~\cite{fujimaki2005approach,ide2004eigenspace}
and disease outbreak detection~\cite{wong2003bayesian}.

Anomalies, which are also called outliers,
are instances that rarely occur.
Therefore, it is natural to consider that instances at a low probability density region are anomalous,
and many density estimation based anomaly detection methods
have been proposed~\cite{barnett1974outliers,parra1996statistical,yeung2003host}.
By the recent advances of deep learning,
the density estimation performance has been greatly improved
by neural network based density estimators,
such as
variational autoencoders (VAE)~\cite{kingma2013auto},
flow-based generative models~\cite{dinh2014nice,dinh2016density,kingma2018glow},
and autoregressive models~\cite{uria2013rnade,raiko2014iterative,germain2015made,uria2016neural}.
The VAE has been used for anomaly detection~\cite{an2015variational,suh2016echo,xu2018unsupervised}.

In some situations, the label information, which indicates
whether each instance is anomalous or normal, is available~\cite{gornitz2013toward}.
The label information is valuable for improving the anomaly detection performance.
However, the existing neural network based density estimation methods cannot
exploit the label information.
To use the anomaly label information,
supervised classifiers, such as
nearest neighbor methods~\cite{singh2009ensemble},
support vector machines~\cite{mukkamala2005model}, and
feed-forward neural networks~\cite{rapaka2003intrusion}, have been used.
However, these standard supervised classifiers do not perform well
when labeled anomalous instances are very few,
which is often the case,
since anomalous instances rarely occur by definition.

In this paper, we propose a neural network density estimator based
anomaly detection method that can exploit the label information.
The proposed method performs well
even when only a few labeled anomalous instances are given
since it is based on a density estimator, which works
without any labeled anomalous instances.
We employ the negative log probability of an instance
as its anomaly score.
For the density function to calculate the probability,
we use neural autoregressive models~\cite{uria2016neural,germain2015made}.
The autoregressive models
can compute the probability density exactly for a test instance.
On the other hand, the VAE computes the lower bound of the probability density approximately.
Moreover, the autoregressive models 
have been achieved the high density estimation performance
compared with other neural density estimators,
such as VAE and flow-based generative models~\cite{dinh2016density}.

The density function is trained
so that the probability density of normal instances becomes high,
which is the same with the standard maximum likelihood estimation.
In addition, we would like to make the density function to satisfy that 
the probability density of anomalous instances is lower than
that of normal instances.
To achieve this, we introduce a regularization term, which is calculated by using
the log likelihood ratio between normal and anomalous instances.
Since our objective function is differentiable,
the density function can be estimated efficiently by using
stochastic gradient-based optimization methods.

\begin{figure*}[t!]
 \centering
\includegraphics[width=30em]{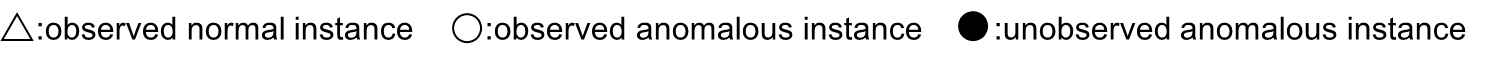}\\
{\tabcolsep=0.3em
 \begin{tabular}{ccc}
\includegraphics[width=16.5em]{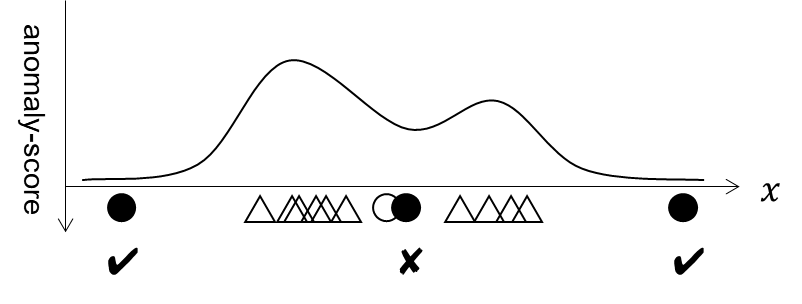}&
\includegraphics[width=16.5em]{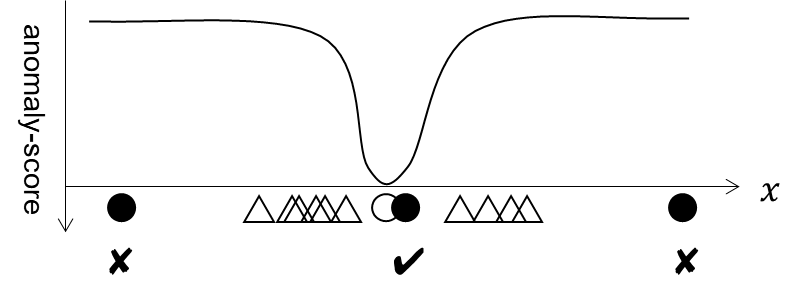}&
\includegraphics[width=16.5em]{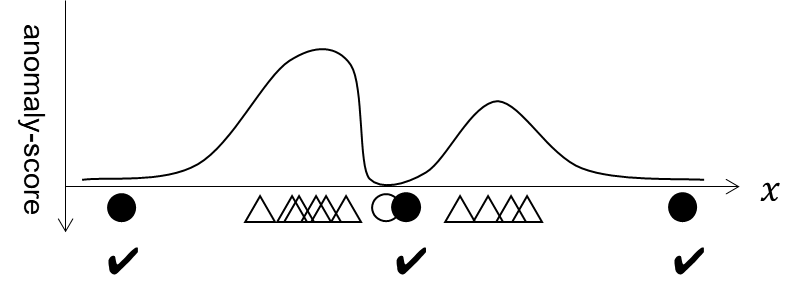}\\
(a) Unsupervised anomaly detection&
(b) Supervised anomaly detection&
(c) Proposed method\\
 \end{tabular}}
\caption{Examples of anomaly scores
with an unsupervised density estimation based anomaly detection method (a),
with a supervised binary classifier based anomaly detection method (b),
and with the proposed method (c).
The white triangle $\vartriangle$ represents an observed normal instance,
the white circle $\circ$ represents an observed anomalous instance,
and the black circle $\bullet$ represents a test anomalous instance, which is not observed in training.
The horizontal axis is the one-dimensional attribute space,
and the vertical axis is the anomaly score, where the anomaly score increases in the downward direction.
The marks $\surd$ and $\times$ indicate that the method can detect the test anomalous instance successfully or not,
respectively.}
\label{fig:example}
\end{figure*}

Figure~\ref{fig:example} illustrates anomaly scores
with an unsupervised density estimation based anomaly detection method (a),
with a supervised binary classifier based anomaly detection method (b),
and with the proposed method (c).
The unsupervised method considers only normal instances, 
and the anomaly score is low where normal instances are located.
Since it cannot exploit the information on anomalous instances,
the anomaly score cannot be increased even where anomalous instances are located closely.
With this example, it succeeds to detect test anomalous instances at the far left and far right,
but fails to detect the test anomalous instance at the center,
where normal instances are closely located.
The supervised method considers both normal and anomalous instances,
where a decision boundary is placed between the normal and anomalous instances.
It can detect the test anomalous instances at the center since an observed anomalous instance exists in the same region.
However, it cannot detect the test anomalous instances at both ends 
since they are at the normal instance side of the decision boundary.
With the proposed method, the anomaly score is high
at the region where normal instances are not located
as well as the region where anomalous instances are located.
Therefore, it can detect all of the test anomalous instances in this example.

The remainder of the paper is organized as follows.
In Section~\ref{sec:proposed}, we define our task,
and propose our method for supervised anomaly detection
based on the neural autoregressive estimators.
In Section~\ref{sec:related}, we briefly review related work.
In Section~\ref{sec:experiments}, we demonstrate the effectiveness
of the proposed method using various datasets.
Finally, we present concluding remarks and
a discussion of future work in Section~\ref{sec:conclusion}.

\section{Proposed method}
\label{sec:proposed}

\subsection{Task}
Suppose that we have a dataset
$\vec{X}=\{(\vec{x}_{n},y_{n})\}_{n=1}^{N}$,
where $\vec{x}_{n}=(x_{n1},\cdots,x_{nD})$ is the $D$-dimensional attribute vector of the $n$th instance,
and $y_{n}$ is its anomaly label, i.e. $y_{n}=1$ if it is anomalous and $y_{n}=0$ if it is not anomalous, or normal.
Our task is to estimate the anomaly score of unseen instances $\vec{x}^{*}$,
where the anomaly score of anomalous instances is high,
and that of normal instances is low.

\subsection{Anomaly score}
The anomalous instances rarely occur, and
the normal instances frequently occur.
Then,
the proposed method uses the following negative log probability
as the anomaly score of instance $\vec{x}$,
\begin{align}
\text{anomaly-score}(\vec{x})=-\log p(\vec{x}|\bm{\theta}),
\end{align}
where $\bm{\theta}$ is parameters of the density function.

\subsection{Density model}

For the density function $p(\vec{x}|\bm{\theta})$,
we use the deep masked autoencoder density estimator (MADE)~\cite{germain2015made},
which is a neural autoregressive model.
The probability distribution
can always be decomposed into the product of
the nested conditional distributions using the probability product rule
as follows,
\begin{align}
p(\vec{x}|\bm{\theta})=\prod_{d=1}^{D}p(x_{d}|\vec{x}_{<d};\bm{\theta}),
\end{align}
where $\vec{x}_{<d}=[x_{1},\cdots,x_{d-1}]$ is the
attribute vector before the $d$th attribute.

We model the conditional distribution with
the following Gaussian mixture,
\begin{align}
p(x_{d}|\vec{x}_{<d};\bm{\theta})=&\sum_{k=1}^{K}\biggl(\pi_{dk}(\vec{x}_{<d};\bm{\theta})
\nonumber\\
&\times\mathcal{N}\left(x_{d}|\mu_{dk}(\vec{x}_{<d};\bm{\theta}),\sigma_{dk}^{2}(\vec{x}_{<d};\bm{\theta})\right)\biggr),
\end{align}
where $K$ is the number of mixture components,
$\mathcal{N}(\cdot|\mu,\sigma^{2})$ is the Gaussian distribution
with mean $\mu$ and variance $\sigma^{2}$,
and $\pi_{dk}(\vec{x}_{<d};\bm{\theta})$, $\mu_{dk}(\vec{x}_{<d};\bm{\theta})$, $\sigma_{dk}^{2}(\vec{x}_{<d};\bm{\theta})$ are the neural networks that define the mixture weight, mean and variance of the $k$th mixture component for the $d$th attribute, respectively,
$\pi_{dk}(\vec{x}_{<d};\bm{\theta})\geq 0$, $\sum_{k=1}^{K}\pi_{dk}(\vec{x}_{<d};\bm{\theta})=1$,
$\sigma_{dk}^{2}(\vec{x}_{<d};\bm{\theta})>0$.

When the feature vector is a binary,
we use the following Bernoulli distribution,
\begin{align}
p(x_{d}|\vec{x}_{<d};\bm{\theta})=\phi_{d}(\vec{x}_{<d};\bm{\theta})^{x_{d}}
(1-\phi_{d}(\vec{x}_{<d};\bm{\theta}))^{1-x_{d}},
\end{align}
where $\phi_{d}(\vec{x}_{<d};\bm{\theta})$
is the neural network that outputs the probability of $x_{d}$ being one.
Similarly, Poisson and Gamma distributions with parameters modeled by
neural networks can be used
in the cases of non-negative integers and non-negative continuous values,
respectively.

With the deep MADE,
the conditional densities of different attributes
are defined by deep autoencoders with masks so that
the conditional density function for the $d$th attribute $x_{d}$
depends only on the attributes before $d$, $\vec{x}_{<d}$, 
but does not depend on the other attributes, $\vec{x}_{\geq d}=[x_{d},\cdots,x_{D}]$.
The MADE is more efficient than other autoregressive models.

Note that in our framework we can use other density estimators, such as VAE and flow-based generative models,
as well as autoencoders, where the reconstruction error is used for the anomaly score.

\subsection{Objective function}
Let $\mathcal{D}=\{1,\cdots,N\}$ be a set of indexes of all the given instances,
$\mathcal{A}=\{n\in\mathcal{D}|y_{n}=1\}$ be a set of indexes of
anomalous instances,
and $\bar{\mathcal{A}}=\{n\in\mathcal{D}|y_{n}=0\}$ be a set of indexes of
normal instances.
The anomaly score of anomalous instances
should be higher than those of normal instances
as follows,
\begin{align}
-\log p(\vec{x}_{n}|\bm{\theta}) > -\log p(\vec{x}_{n'}|\bm{\theta}),
\quad \text{for $n\in\mathcal{A}$, $n'\in\bar{\mathcal{A}}$}.
\label{eq:constraint}
\end{align}
In addition, the following log likelihood of the normal instances should be high,
\begin{align}
L'(\bm{\theta}) = \frac{1}{|\bar{\mathcal{A}}|}
\sum_{n\in\bar{\mathcal{A}}} \log p(\vec{x}_{n}|\bm{\theta}),
\label{eq:likelihood}
\end{align}
since the anomaly score, which is defined by the negative log likelihood,
of the normal instances should be low.
Here, $|\cdot|$ represents the number of elements in the set.

\begin{figure}
  \centering
 \includegraphics[width=23em]{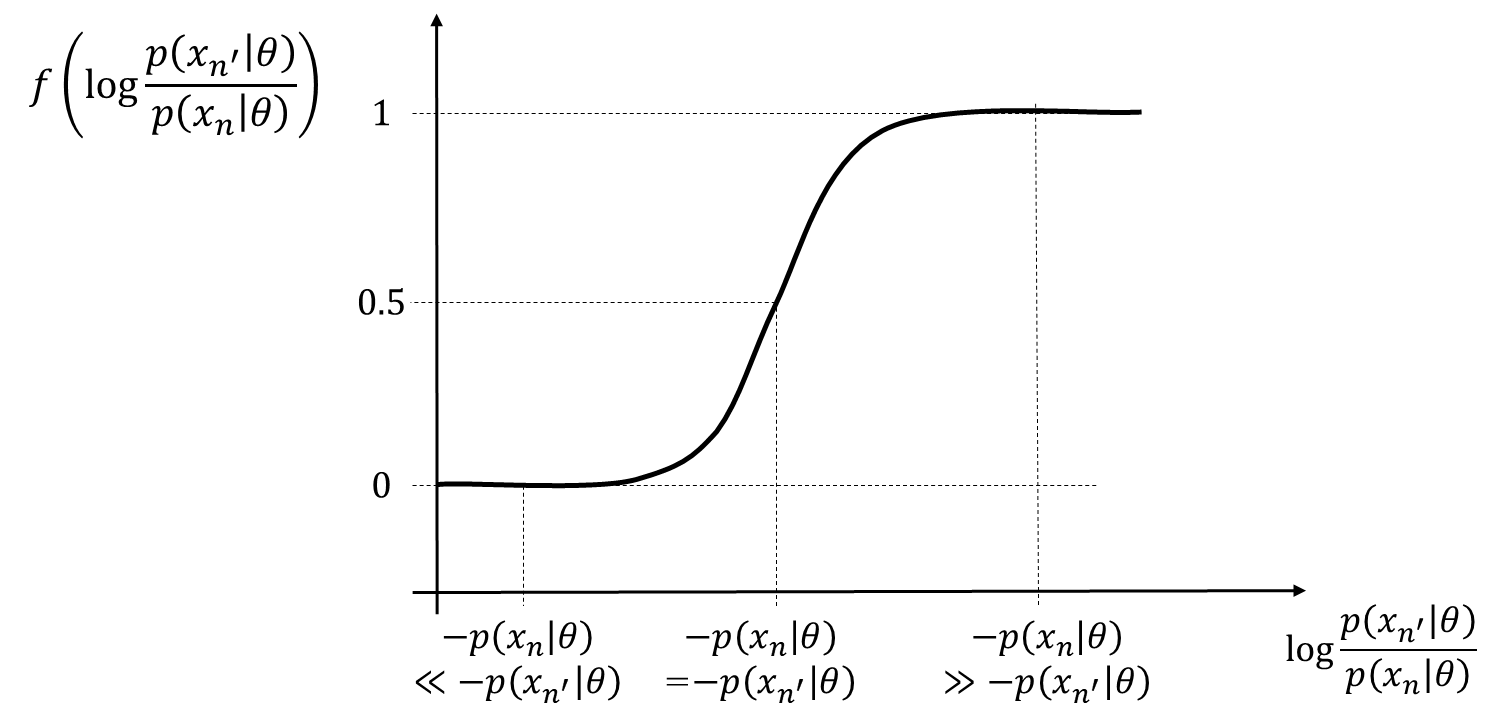}
 \caption{Regularization term $f\left(\log\frac{p(\vec{x}_{n'}|\bm{\theta})}{p(\vec{x}_{n}|\bm{\theta})}\right)$
 of the objective function (\ref{eq:objective}) 
 with respect to the log likelihood ratio $\log\frac{p(\vec{x}_{n'}|\bm{\theta})}{p(\vec{x}_{n}|\bm{\theta})}$.}
 \label{fig:sigmoid}
 \end{figure}

We would like to maximize Eq. (\ref{eq:likelihood}) while satisfying the
constraints in Eq. (\ref{eq:constraint}) as much as possible.
To make the objective function differentiable with respect to the parameters
without constraints,
we relax the constraints in Eq. (\ref{eq:constraint})
to a soft regularization term as follows,
\begin{align}
L(\bm{\theta}) = L'(\bm{\theta})+
\frac{\lambda}{|\mathcal{A}||\bar{\mathcal{A}}|}
\sum_{n\in\mathcal{A}}\sum_{n'\in\bar{\mathcal{A}}}f\left(
\log\frac{p(\vec{x}_{n'}|\bm{\theta})}{p(\vec{x}_{n}|\bm{\theta})}\right),
\label{eq:objective}
\end{align}
where $f(\cdot)$ is the sigmoid function,
\begin{align}
f(s)=\frac{1}{1+\exp(-s)},
\end{align}
and $\lambda\geq 0$ is the hyperparameter.
Figure~\ref{fig:sigmoid} shows the regularization
with respect to the log likelihood ratio.
When the anomaly score of an anomalous instance
is much higher than that of a normal instance,
$-\log p(\vec{x}_{n}|\bm{\theta}) \gg -\log p(\vec{x}_{n'}|\bm{\theta})$,
the sigmoid function takes the maximum value one.
When the anomaly score of an anomalous instance
is much lower than that of a normal instance,
$-\log p(\vec{x}_{n}|\bm{\theta}) \ll -\log p(\vec{x}_{n'}|\bm{\theta})$,
the sigmoid function takes the minimum value zero.
Therefore, the maximization of this regularization term
moves parameters $\bm{\theta}$ so as to satisfy the constraints in Eq. (\ref{eq:constraint}).
We maximize the objective function (\ref{eq:objective})
with a gradient based optimization method,
such as ADAM~\cite{kingma2014adam}.

When there are no labeled anomalous instances or $\lambda=0$,
the regularization term becomes zero, and the first term on the likelihood remains with the objective function,
which is the same objective function with the standard density estimation.
Therefore, the proposed method can be seen as
a generalization of unsupervised density estimation based anomaly detection methods.

The regularization term can be seen as a stochastic version
of the area under the receiver operating characteristic (ROC) curve (AUC)~\cite{yan2003optimizing},
since the AUC is computed by
\begin{align}
{\rm AUC}
&= \frac{1}{|\mathcal{A}||\bar{\mathcal{A}}|}
\sum_{n\in\mathcal{A}}\sum_{n'\in\bar{\mathcal{A}}}
I(-\log p(\vec{x}_{n}|\bm{\theta}) > -\log p(\vec{x}_{n'}|\bm{\theta}))
\nonumber\\
&= \frac{1}{|\mathcal{A}||\bar{\mathcal{A}}|}
\sum_{n\in\mathcal{A}}\sum_{n'\in\bar{\mathcal{A}}}
I\left(\log \frac{p(\vec{x}_{n'}|\bm{\theta})}{p(\vec{x}_{n}|\bm{\theta})}>0\right)
\end{align}
where $I(A)$ is the indicator function,
i.e. $I(A)=1$ if $A$ is true, $I(A)=0$ otherwise,
and the sigmoid function
is an approximation of the indicator function $f(s)\approx I(s>0)$.

\section{Related work}
\label{sec:related}

A number of unsupervised methods for anomaly detection,
which is sometimes called outlier detection~\cite{hodge2004survey} 
or novelty detection~\cite{markou2003novelty},
have been proposed,
such as the local outlier factor~\cite{breunig2000lof},
one-class support vector machines~\cite{scholkopf2001estimating},
and the isolation forest~\cite{liu2008isolation}.
With density estimation based anomaly detection methods,
Gaussian distributions~\cite{shewhart1931economic},
Gaussian mixtures~\cite{eskin2000anomaly} and kernel density estimators~\cite{laxhammar2009anomaly}
have been used.
The density estimation methods have been regarded as unsuitable
for anomaly detection in high-dimensional data due to
the difficulty of estimating
multivariate probability distributions~\cite{friedland2014classifier,hido2011statistical}.
Although some supervised anomaly detection methods have been 
proposed~\cite{nadeem2016semi,gao2006novel,das2016incorporating,das2017incorporating,Munawar_2017,pimentel2018generalized,akcay2018ganomaly,yamanaka2019autoencoding},
they are not based on deep autoregressive density estimators,
which can achieve high density estimation performance.

Recent research of neural networks has made substantial progress
on density estimation for high-dimensional data.
The neural network based density estimators,
including VAE~\cite{kingma2013auto},
flow-based generative models~\cite{dinh2014nice,dinh2016density,kingma2018glow},
and autoregressive models~\cite{uria2013rnade,raiko2014iterative,germain2015made,uria2016neural},
can flexibly learn dependencies across different attributes,
and have achieved the high density estimation performance.
The autoregressive models have been successfully used for density estimations,
as well as modeling images~\cite{oord2016pixel} and speech~\cite{van2016wavenet}.
The autoregressive models can compute the probability for each instance exactly,
which is desirable since we use the probability as the anomaly score.
A shortcoming of the autoregressive models is that it requires computational time
to generate samples. However, generating samples is not necessary for anomaly detection.
The VAE can compute the approximation of the lower bound of the log probability,
but it cannot compute the probability exactly.
By using importance sampling~\cite{burda2015importance},
one can calculate a lower bound that approaches the true log probability
as the number of samples increases, although
it requires the infinite number of samples for the true probability.
The flow-based generative models can calculate the probability exactly,
and can generate samples efficiently and effectively~\cite{kingma2018glow}.
However, it requires specialized functions for transformations,
which are invertible and their determinant of the Jacobian matrix can be easily calculated,
and the density estimation performance is lower than the autoregressive models.
Although we use the autoregressive models in our framework,
VAE and flow-based generative models are straightforwardly applicable
to our framework.
The VAE has been used for unsupervised anomaly detection~\cite{an2015variational,suh2016echo,xu2018unsupervised},
but not for supervised anomaly detection.

In the proposed method with the high value of the hyperparameter $\lambda$,
the second term in the objective function (\ref{eq:objective}),
which is an approximation of the AUC, is dominant.
Therefore, the proposed method is related to AUC maximization~\cite{cortes2004auc,brefeld2005auc},
which has been used for training on class imbalanced data.
The proposed method employs the likelihood maximization with normal instances
as well as the AUC maximization,
which enables us to improve the performance with a few training anomalous instances.
With our experiments described in Section~\ref{sec:experiments},
we demonstrate that both of the likelihood and AUC maximizations are effective
to achieve good performance in various datasets.

\section{Experiments}
\label{sec:experiments}

\subsection{Data}

We evaluated our proposed supervised anomaly detection method based on deep autoregressive density estimators
with 16 datasets used for unsupervised outlier detection~\cite{campos2016evaluation}~\footnote{The datasets were obtained from \url{http://www.dbs.ifi.lmu.de/research/outlier-evaluation/DAMI/}}.
The number of instances $N$,
the number of attributes $D$,
the number of anomalous instances $|\mathcal{A}|$
and the anomaly rate $|\mathcal{A}|/N$ of the datasets
are shown in Table~\ref{tab:stats}.
Each attribute was linearly normalized to the range $[0,1]$,
and duplicate instances were removed.
We used 80\% of the normal instances and three anomalous instances for training,
10\% of the normal instances and three anomalous instances for validation,
and remaining instances for testing.
For the evaluation measurement, we used the AUC. 
For each dataset,
we randomly generated ten sets of training, validation and test data,
and calculated the average AUC over the ten sets.

\begin{table}
\centering
\caption{Statistics of the datasets used in our experiments. $N$ is the number of instances,
$D$ is the number of attributes,
$|\mathcal{A}|$ is the number of anomalous instances, 
and $|\mathcal{A}|/N$ is the anomaly rate.}
\label{tab:stats}
\begin{tabular}{lrrrr}
\hline
data & $N$ & $D$ & $|\mathcal{A}|$ & $|\mathcal{A}|/N$\\
\hline
Annthyroid & 7016 & 21 & 350 & 0.050 \\
Arrhythmia & 305 & 259 & 61 & 0.200 \\
Cardiotocography & 2068 & 21 & 413 & 0.200 \\
HeartDisease & 187 & 13 & 37 & 0.198 \\
InternetAds & 1775 & 1555 & 177 & 0.100 \\
Ionosphere & 351 & 32 & 126 & 0.359 \\
KDDCup99 & 60839 & 79 & 246 & 0.004 \\
PageBlocks & 5171 & 10 & 258 & 0.050 \\
Parkinson & 60 & 22 & 12 & 0.200 \\
PenDigits & 9868 & 16 & 20 & 0.002 \\
Pima & 625 & 8 & 125 & 0.200 \\
Shuttle & 1013 & 9 & 13 & 0.013 \\
SpamBase & 3485 & 57 & 697 & 0.200 \\
Stamps & 325 & 9 & 16 & 0.049 \\
Waveform & 3443 & 21 & 100 & 0.029 \\
Wilt & 4671 & 5 & 93 & 0.020 \\
\hline
\end{tabular}
\end{table}

\subsection{Comparing methods}

We compared the proposed method with the following nine methods:
LOF, OCSVM, IF, VAE, MADE, KNN, SVM, RF and NN.
LOF, OCSVM, IF, VAE and MADE are unsupervised anomaly detection methods, 
where the attribute $\vec{x}_{n}$ is used for calculating the anomaly score,
but the label information $y_{n}$ is not used.
KNN, SVM, RF, NN as well as the proposed method are supervised anomaly detection methods,
where both the attribute $\vec{x}_{n}$ and the label information $y_{n}$ are used.
The hyperparameters were selected based on the AUC score on the validation data for both of the unsupervised
and supervised methods.
We used the implementation of scikit-learn~\cite{pedregosa2011scikit} with 
LOF, OCSVM, IF, KNN, SVM, RF and NN.

\begin{itemize}
\item {\bf LOF} is the local outlier factor method~\cite{breunig2000lof}.
The LOF unsupervisedly detects anomalies
based on the degree of isolation from the surrounding neighborhood.
The number of neighbors was tuned from $\{1,3,5,15,35\}$ using the validation data.

\item {\bf OCSVM} is the one-class support vector machine~\cite{scholkopf2001estimating},
which is an extension of the support vector machine (SVM) to the case of unlabeled data.
The OCSVM finds the maximal margin hyperplane which 
separates the given normal data from the origin by embedding them into a high dimensional space via a kernel function.
We used the RBF kernel,
and the kernel hyperparameter was tuned from $\{10^{-3},10^{-2},10^{-1},1\}$.

\item {\bf IF} is the isolation forest method~\cite{liu2008isolation}, which is a tree-based
unsupervised anomaly detection method. The IF isolates anomalies by randomly selecting an attribute
and randomly selecting a split value between the maximum and minimum values of the selected attribute.
The number of base estimators was chosen from $\{1,5,10,20,30\}$.

\item {\bf VAE} is the variational autoencoder~\cite{kingma2013auto},
which is a density estimation method based on neural networks.
With the VAE, the observation is assumed to follow a Gaussian distribution,
where the mean and variance are modeled by a neural network that takes latent variables as the input.
The latent variable is also modeled by another neural network that takes the
attribute vector as the input.
We used three-layered feed-forward neural networks with 100 hidden units,
and the 20-dimensional latent space.
We optimized the neural network parameters using ADAM.
The number of epochs was selected by using the validation data.

\item {\bf MADE} is the deep masked autoencoder density estimator~\cite{germain2015made},
which is used with the proposed method for the density function.
The proposed method with $\lambda=0$ corresponds to the MADE.
We used the same parameter setting with the proposed method for the MADE,
which is described in the next subsection.

\item {\bf KNN} is the $k$-nearest neighbor method,
which classifies instances based on the votes of the neighbors.
The number of neighbors was selected from $\{1,3,5,15\}$.

\item {\bf SVM} is the support vector machine~\cite{scholkopf2002learning},
which is a kernel-based binary classification method.
We used the RBF kernel,
and the kernel hyperparameter was tuned from $\{10^{-3},10^{-2},10^{-1},1\}$.

\item {\bf RF} is the random forest method~\cite{breiman2001random},
which is a meta estimator that fits a number of decision tree classifiers.
The number of trees was chosen from $\{5,10,20,30\}$.

\item {\bf NN} is the feed-forward neural network classifier.
We used three layers with rectified linear unit (ReLU) activation,
       where the number of hidden units was selected from $\{5,10,50,100\}$.
\end{itemize}

\subsection{Settings of the proposed method}

We used Gaussian mixtures with $K=3$ components for the output layer.
The number of hidden layers was one,
the number of hidden units was 500,
the number of masks was ten, and
the number of different orderings was ten.
The hyperparameter $\lambda$ was selected from $\{0,10^{-1},1,10,10^{2},10^{3},10^{4}\}$
using the validation data.
The validation data were also used for early stopping, where the maximum number of training epochs was 100.
We optimized the neural network parameters using ADAM with learning rate $10^{-3}$.

\subsection{Results}

\begin{table*}
 \centering
 \caption{AUCs on 16 datasets with three training anomalous instances by unsupervised anomaly detection methods (LOF, OCSVM, IF, VAE, MADE) and supervised anomaly detection methods (KNN, SVM, RF, NN, Proposed). Values in bold typeface are not statistically different (at the 5\% level)
 from the best performing method according to a paired t-test. The bottom row shows the average AUC over the datasets.}
 \label{tab:auc3}
\begin{tabular}{lrrrrrrrrrr}
\hline
& LOF & OCSVM & IF & VAE & MADE & KNN & SVM & RF & NN & Proposed \\
\hline
Annthyroid & 0.627 & 0.667 & 0.700 & 0.766 & 0.716 & 0.510 & 0.741 & {\bf 0.875} & 0.596 & 0.776\\
Arrhythmia & {\bf 0.711} & {\bf 0.718} & 0.649 & 0.668 & {\bf 0.694} & 0.504 & 0.568 & 0.621 & 0.638 & {\bf 0.677}\\
Cardiotocography & 0.569 & 0.834 & {\bf 0.836} & 0.726 & {\bf 0.828} & 0.582 & {\bf 0.846} & 0.707 & 0.554 & {\bf 0.873}\\
HeartDisease & 0.581 & 0.688 & 0.759 & 0.729 & {\bf 0.803} & 0.664 & {\bf 0.722} & 0.701 & 0.540 & {\bf 0.825}\\
InternetAds & 0.677 & {\bf 0.836} & 0.562 & {\bf 0.860} & {\bf 0.780} & 0.533 & {\bf 0.804} & 0.601 & 0.711 & {\bf 0.834}\\
Ionosphere & 0.860 & {\bf 0.951} & 0.864 & 0.844 & 0.821 & 0.564 & {\bf 0.933} & 0.883 & 0.792 & 0.845\\
KDDCup99 & 0.582 & 0.993 & 0.992 & {\bf 0.968} & {\bf 0.995} & 0.707 & 0.812 & 0.939 & 0.979 & {\bf 0.988}\\
PageBlocks & 0.776 & {\bf 0.930} & {\bf 0.923} & {\bf 0.924} & 0.847 & 0.560 & 0.730 & 0.675 & 0.428 & 0.815\\
Parkinson & {\bf 0.840} & {\bf 0.847} & {\bf 0.748} & {\bf 0.747} & {\bf 0.797} & 0.640 & {\bf 0.817} & {\bf 0.735} & {\bf 0.733} & {\bf 0.807}\\
PenDigits & 0.898 & 0.989 & 0.955 & 0.915 & 0.901 & 0.839 & {\bf 0.999} & 0.946 & 0.617 & 0.993\\
Pima & 0.586 & {\bf 0.688} & {\bf 0.737} & {\bf 0.677} & {\bf 0.725} & 0.530 & 0.649 & 0.569 & 0.296 & {\bf 0.744}\\
Shuttle & 0.962 & 0.918 & 0.949 & 0.952 & {\bf 0.927} & 0.879 & {\bf 0.997} & {\bf 0.999} & 0.400 & {\bf 0.969}\\
SpamBase & 0.521 & 0.662 & {\bf 0.781} & {\bf 0.775} & {\bf 0.735} & 0.536 & {\bf 0.750} & {\bf 0.766} & 0.621 & {\bf 0.786}\\
Stamps & 0.814 & {\bf 0.890} & {\bf 0.922} & {\bf 0.908} & {\bf 0.902} & 0.761 & {\bf 0.895} & {\bf 0.868} & 0.832 & {\bf 0.904}\\
Waveform & {\bf 0.729} & {\bf 0.737} & 0.709 & {\bf 0.751} & {\bf 0.743} & 0.532 & {\bf 0.801} & 0.591 & 0.709 & {\bf 0.800}\\
Wilt & 0.731 & 0.352 & 0.596 & 0.455 & 0.707 & 0.503 & {\bf 0.801} & 0.623 & 0.649 & {\bf 0.785}\\
\hline
average & 0.717 & 0.794 & 0.793 & 0.791 & 0.807 & 0.615 & 0.804 & 0.756 & 0.631 & {\bf 0.839} \\
\hline
\end{tabular}
\end{table*}

\begin{table*}[t!]
 \centering
 \caption{AUCs on datasets with (a) one and (b) five training anomalous instances by supervised anomaly detection methods. The AUCs by unsupervised methods are the same with Table~\ref{tab:auc3} since they do not use the anomaly information for training. Values in bold typeface are not statistically different (at the 5\% level)  from the best performing method according to a paired t-test including AUCs by the unsupervised methods. The AUCs on Parkinson and Shuttle datasets with five training anomalous instances could not be calculated since they do not contain enough anomalous instances.}
 \label{tab:auc15}
    \begin{minipage}[5]{.50\textwidth}
        \centering
\begin{tabular}{lrrrrr}
& \multicolumn{5}{c}{(a) one training anomalous instance}\\
\hline
& KNN & SVM & RF & NN & Proposed \\
 \hline
 Annthyroid & 0.506 & 0.664 & 0.691 & 0.586 & 0.726\\
Arrhythmia & 0.503 & 0.522 & 0.545 & 0.550 & {\bf 0.680}\\
Cardiotocography & 0.529 & 0.744 & 0.565 & 0.500 & {\bf 0.846}\\
HeartDisease & 0.542 & 0.765 & 0.605 & 0.292 & {\bf 0.820}\\
InternetAds & 0.513 & 0.700 & 0.546 & 0.615 & {\bf 0.851}\\
Ionosphere & 0.516 & {\bf 0.919} & 0.745 & 0.758 & 0.820\\
KDDCup99 & 0.575 & 0.833 & 0.791 & 0.979 & {\bf 0.949}\\
PageBlocks & 0.514 & 0.687 & 0.530 & 0.421 & 0.848\\
Parkinson & 0.642 & {\bf 0.847} & {\bf 0.762} & 0.445 & {\bf 0.850}\\
PenDigits & 0.766 & {\bf 0.996} & 0.648 & 0.483 & 0.987\\
Pima & 0.506 & 0.586 & 0.517 & 0.365 & 0.691\\
Shuttle & 0.661 & {\bf 0.989} & {\bf 0.861} & 0.532 & {\bf 0.985}\\
SpamBase & 0.510 & {\bf 0.758} & 0.697 & 0.462 & 0.703\\
Stamps & 0.639 & 0.826 & 0.726 & 0.795 & 0.849\\
Waveform & 0.506 & {\bf 0.752} & 0.510 & 0.695 & {\bf 0.748}\\
 Wilt & 0.503 & {\bf 0.725} & 0.543 & 0.614 & {\bf 0.776}\\
 \hline
average  & 0.558 & 0.770 & 0.643 & 0.568 & {\bf 0.821} \\
 \hline
\end{tabular}
\end{minipage}
\hfill
\begin{minipage}[t]{.40\textwidth}
\centering
 \begin{tabular}{rrrrr}
\multicolumn{5}{c}{(b) five training anomalous instances}\\
\hline
KNN & SVM & RF & NN & Proposed \\
\hline
0.517 & {\bf 0.807} & {\bf 0.857} & 0.610 & 0.761\\
0.513 & {\bf 0.650} & 0.642 & 0.657 & {\bf 0.701}\\
0.589 & {\bf 0.885} & 0.790 & 0.594 & {\bf 0.901}\\
0.740 & {\bf 0.809} & 0.779 & 0.695 & {\bf 0.824}\\
0.568 & 0.849 & 0.666 & 0.786 & 0.866\\
0.713 & {\bf 0.943} & {\bf 0.945} & 0.817 & 0.887\\
0.754 & 0.890 & 0.962 & 0.980 & {\bf 0.979}\\
0.630 & 0.726 & 0.789 & 0.450 & 0.891\\
 \\
0.890 & {\bf 0.999} & {\bf 0.969} & 0.733 & 0.995\\
0.546 & 0.663 & 0.653 & 0.304 & {\bf 0.771}\\
 \\
0.562 & 0.814 & {\bf 0.869} & 0.780 & {\bf 0.852}\\
0.873 & {\bf 0.946} & {\bf 0.924} & 0.805 & {\bf 0.920}\\
0.560 & {\bf 0.894} & 0.678 & 0.750 & {\bf 0.866}\\
0.512 & {\bf 0.852} & 0.730 & 0.654 & 0.812\\
 \hline
0.640 & 0.838 & 0.804 & 0.687 & {\bf 0.859} \\
\hline		
 \end{tabular}
 \end{minipage}
 \end{table*}

   \begin{figure*}[t!]
    \centering
   {\tabcolsep=0.0em
   \begin{tabular}{cccccc}
    \includegraphics[width=8.3em]{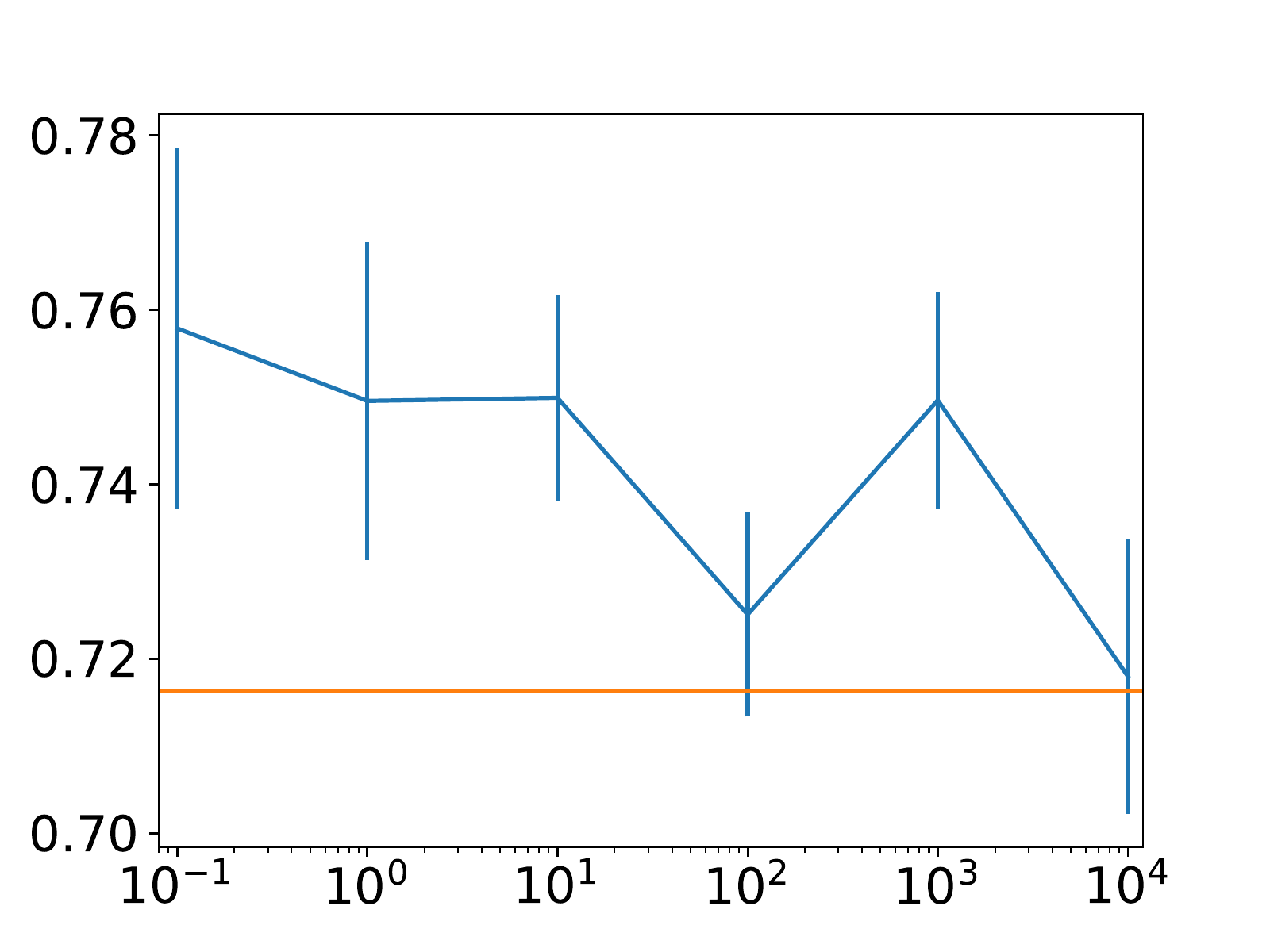} &
\includegraphics[width=8.3em]{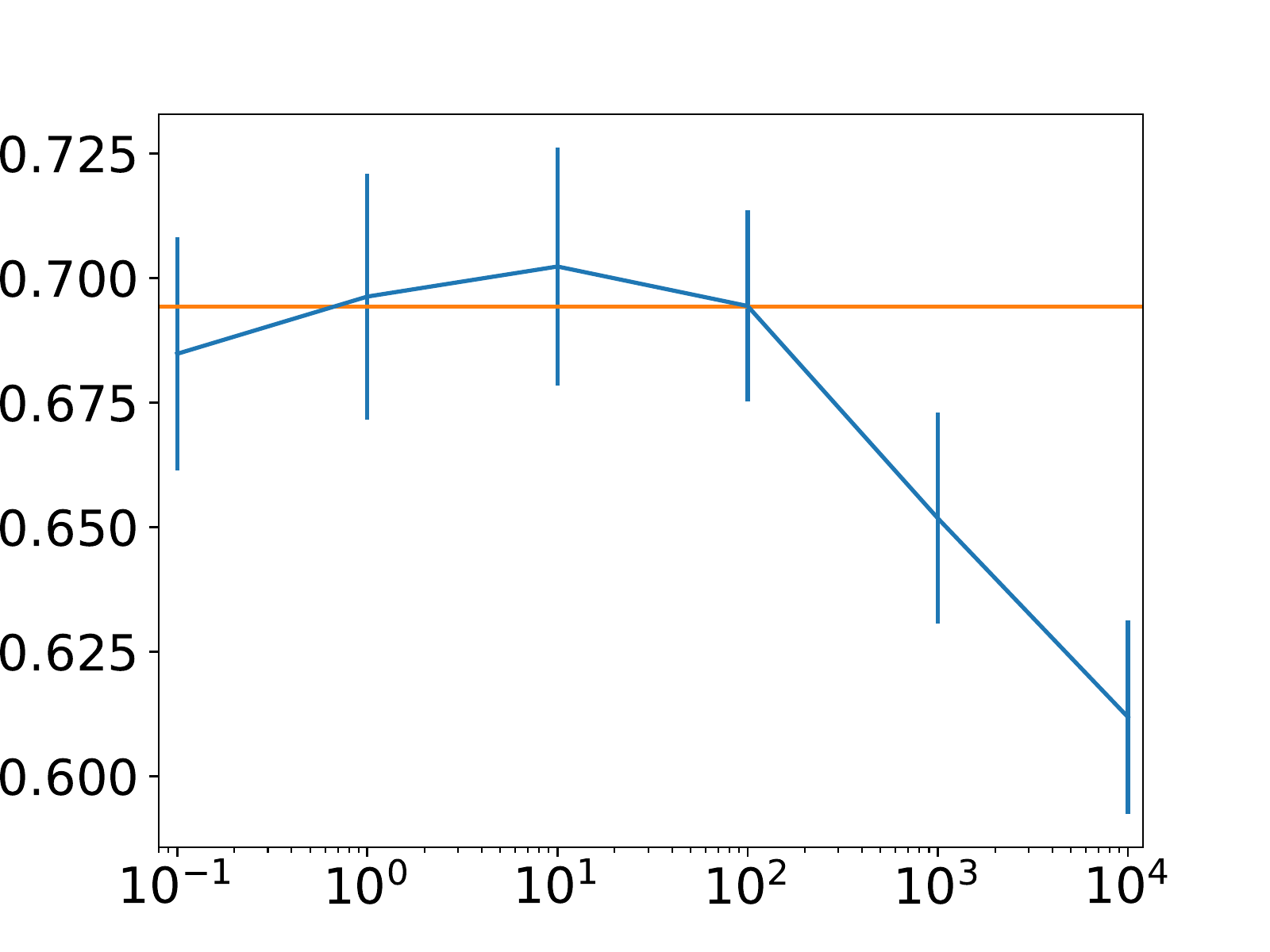} &
\includegraphics[width=8.3em]{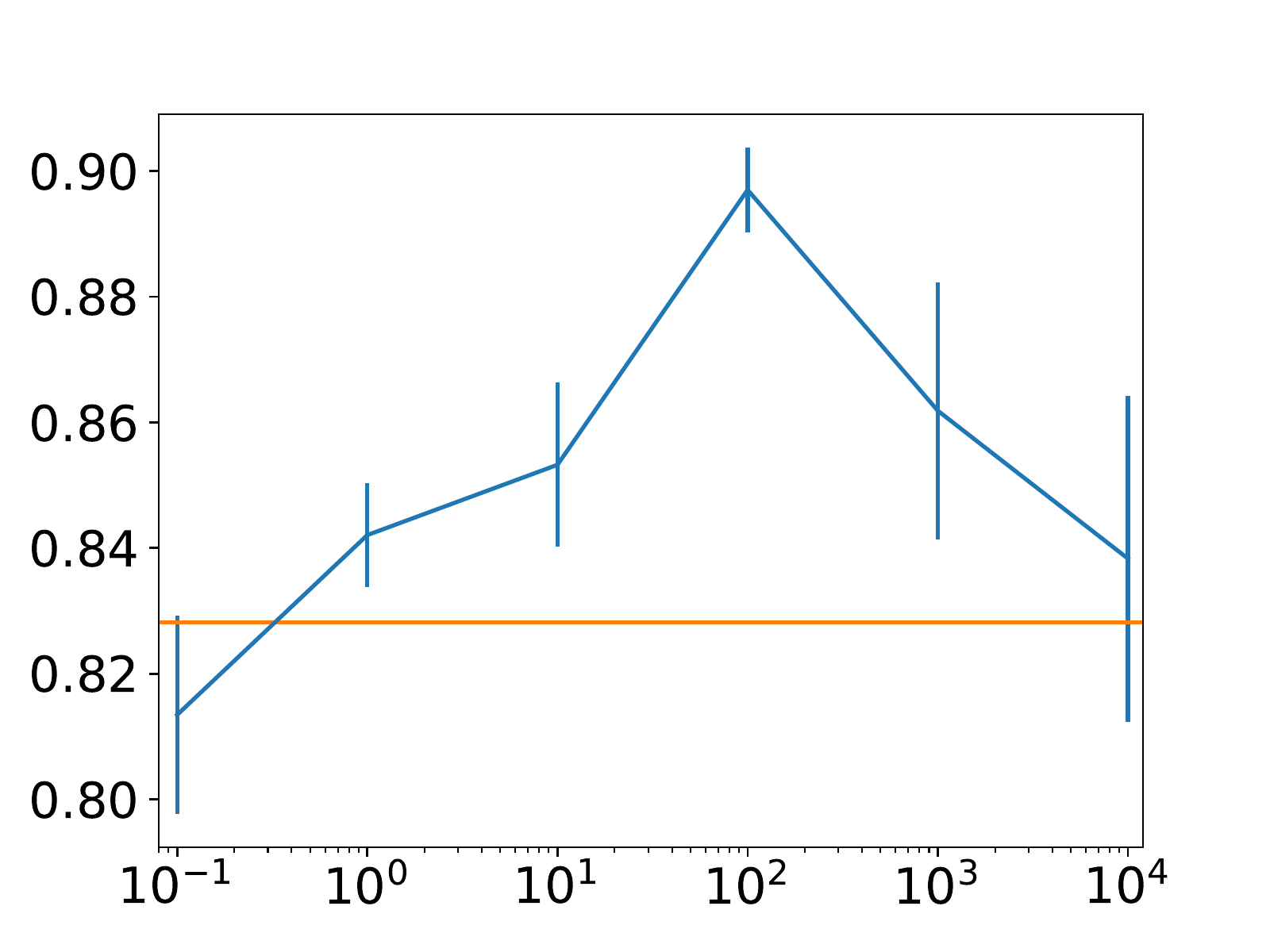} &
\includegraphics[width=8.3em]{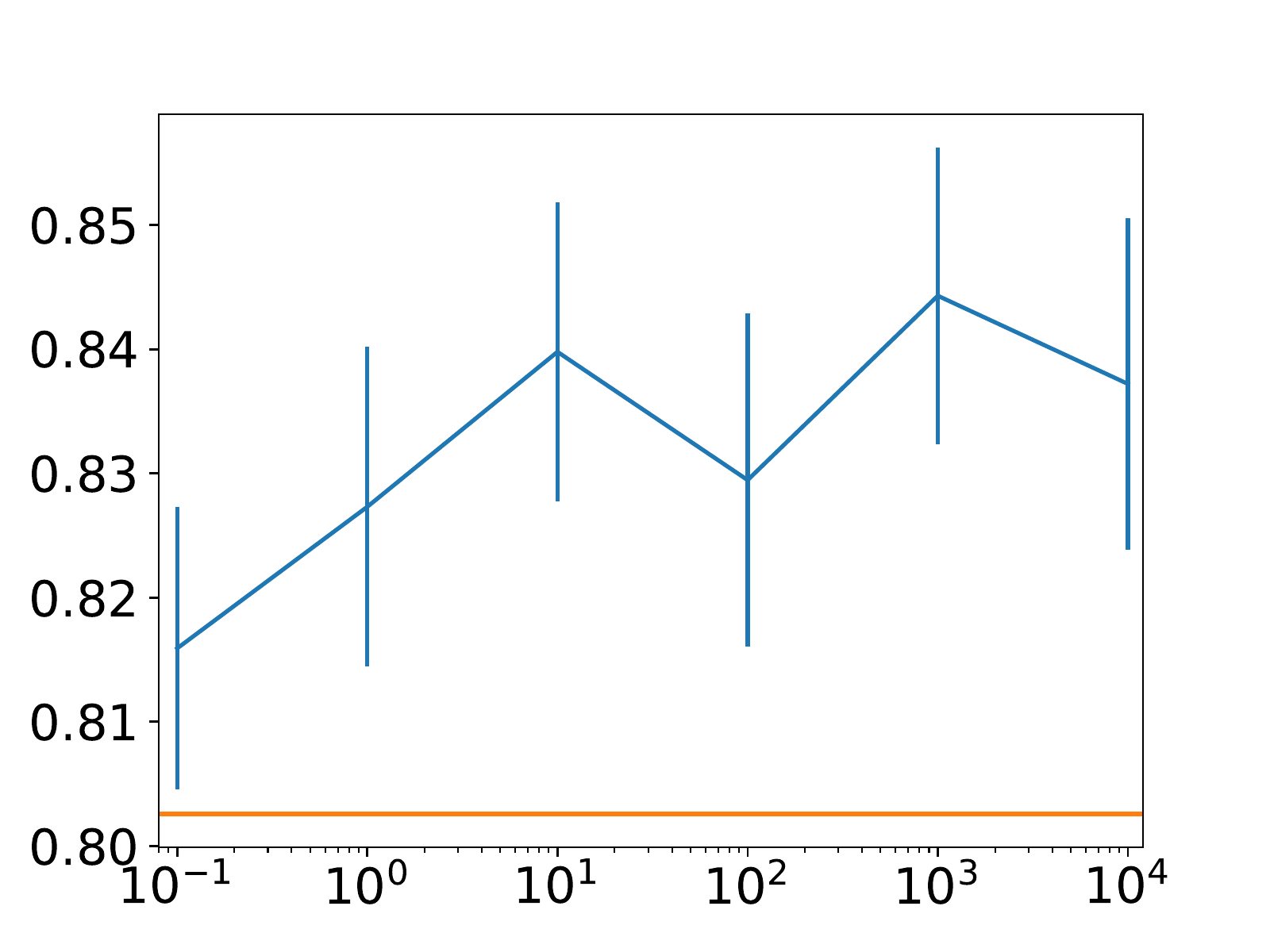} &
\includegraphics[width=8.3em]{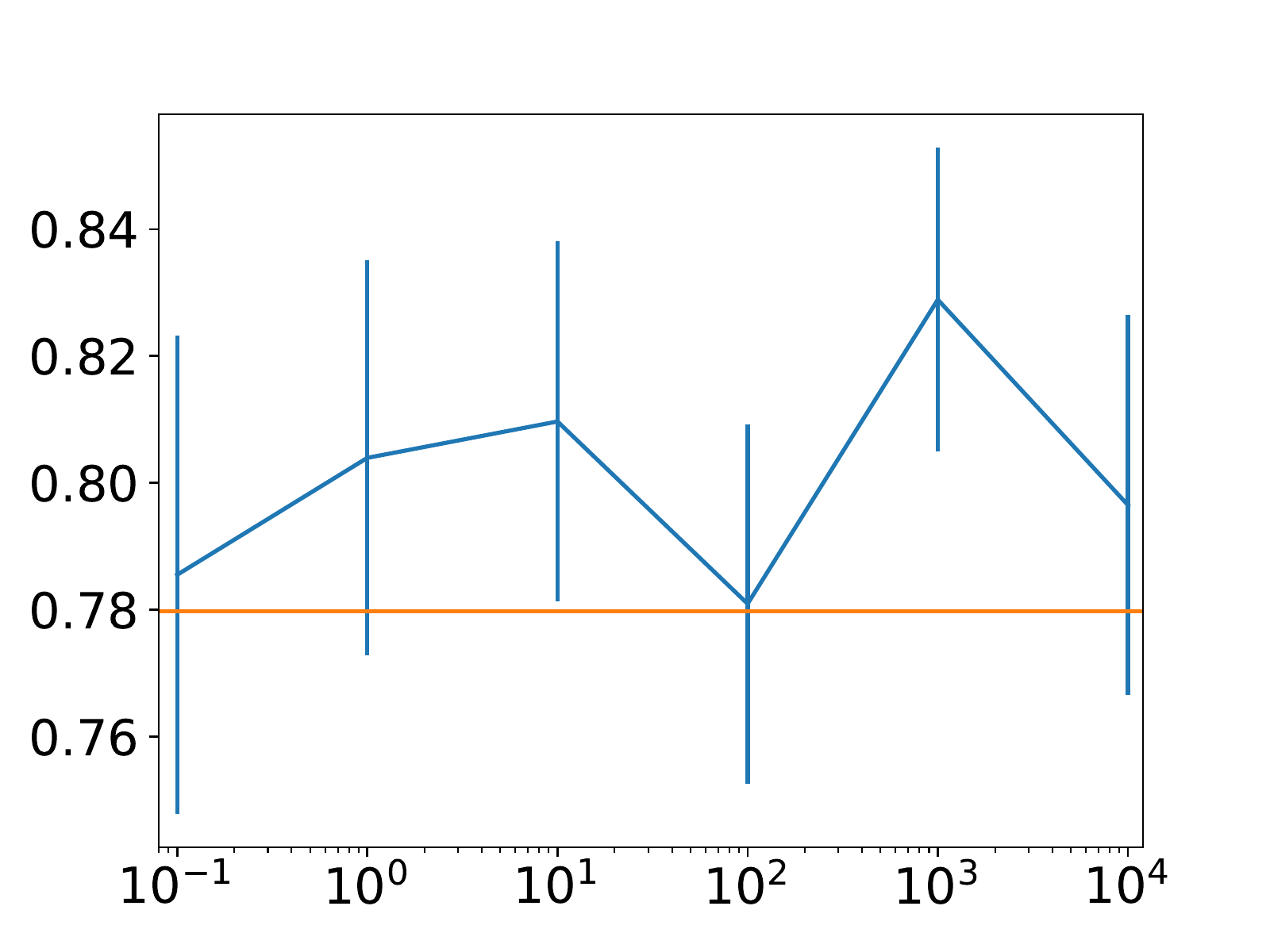} &
    \includegraphics[width=8.3em]{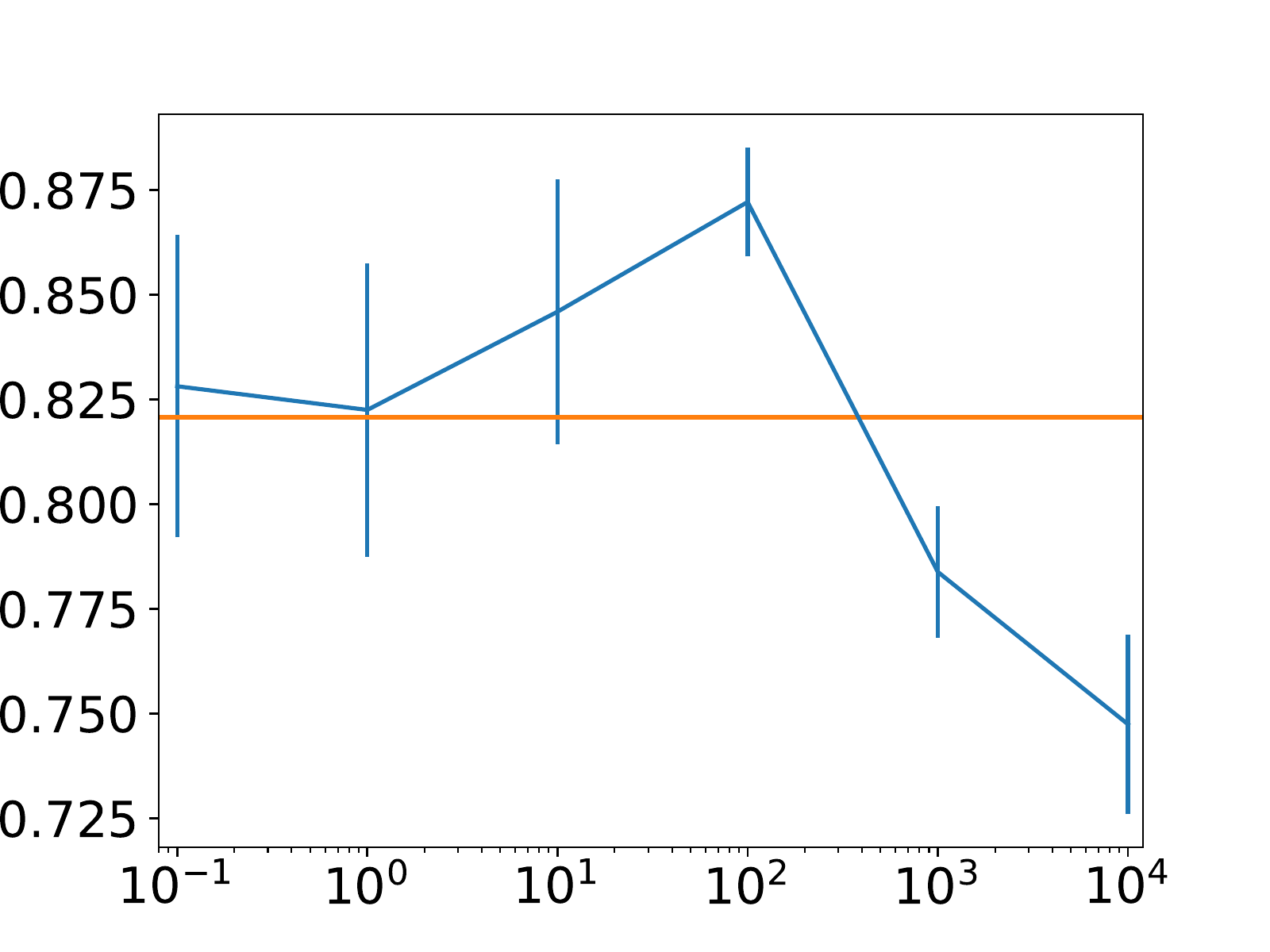} \\
Annthyroid&
Arrhythmia&
Cardiotocography&
HeartDisease&
InternetAds&
Ionosphere\\
\includegraphics[width=8.3em]{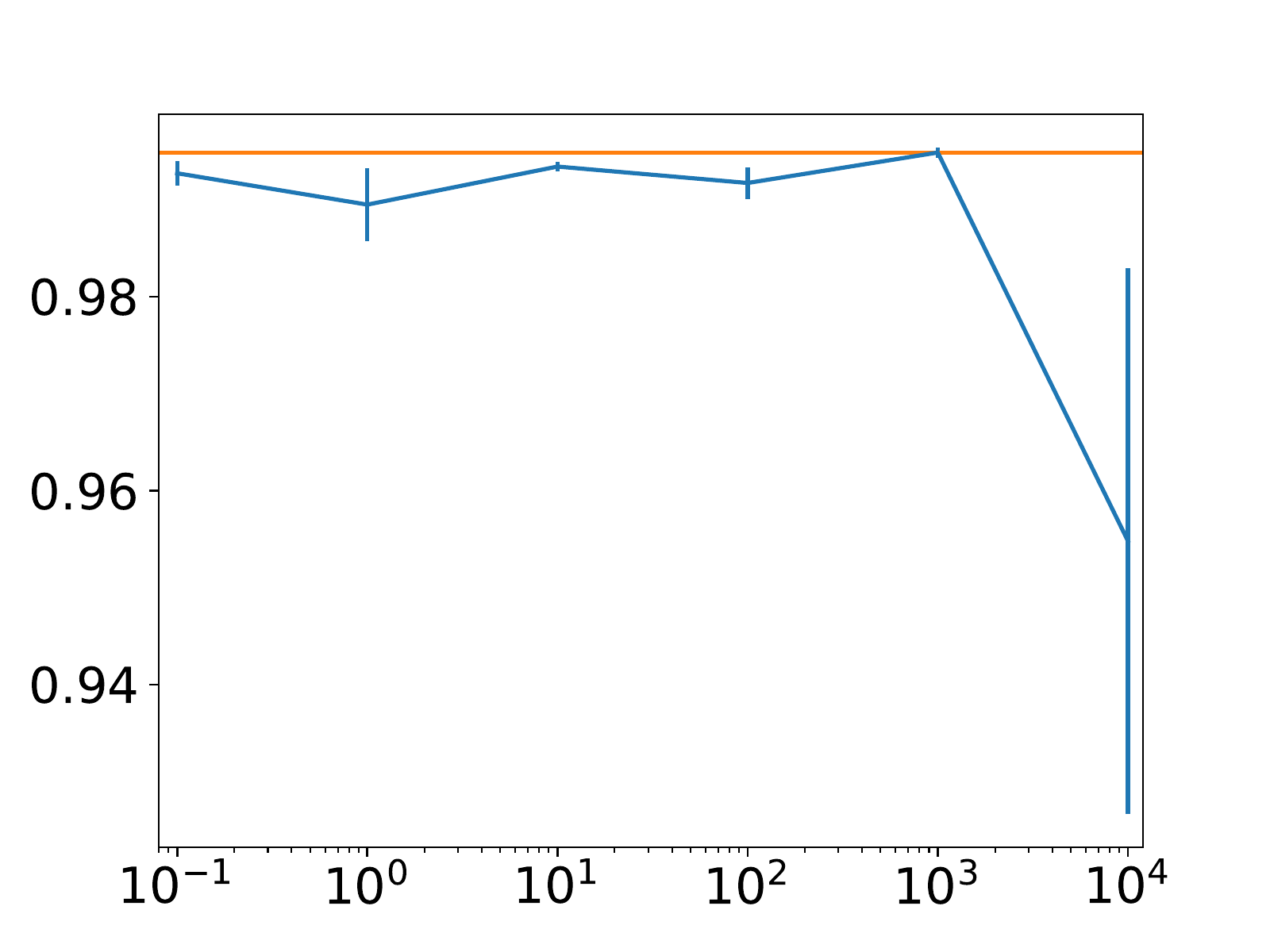} &
\includegraphics[width=8.3em]{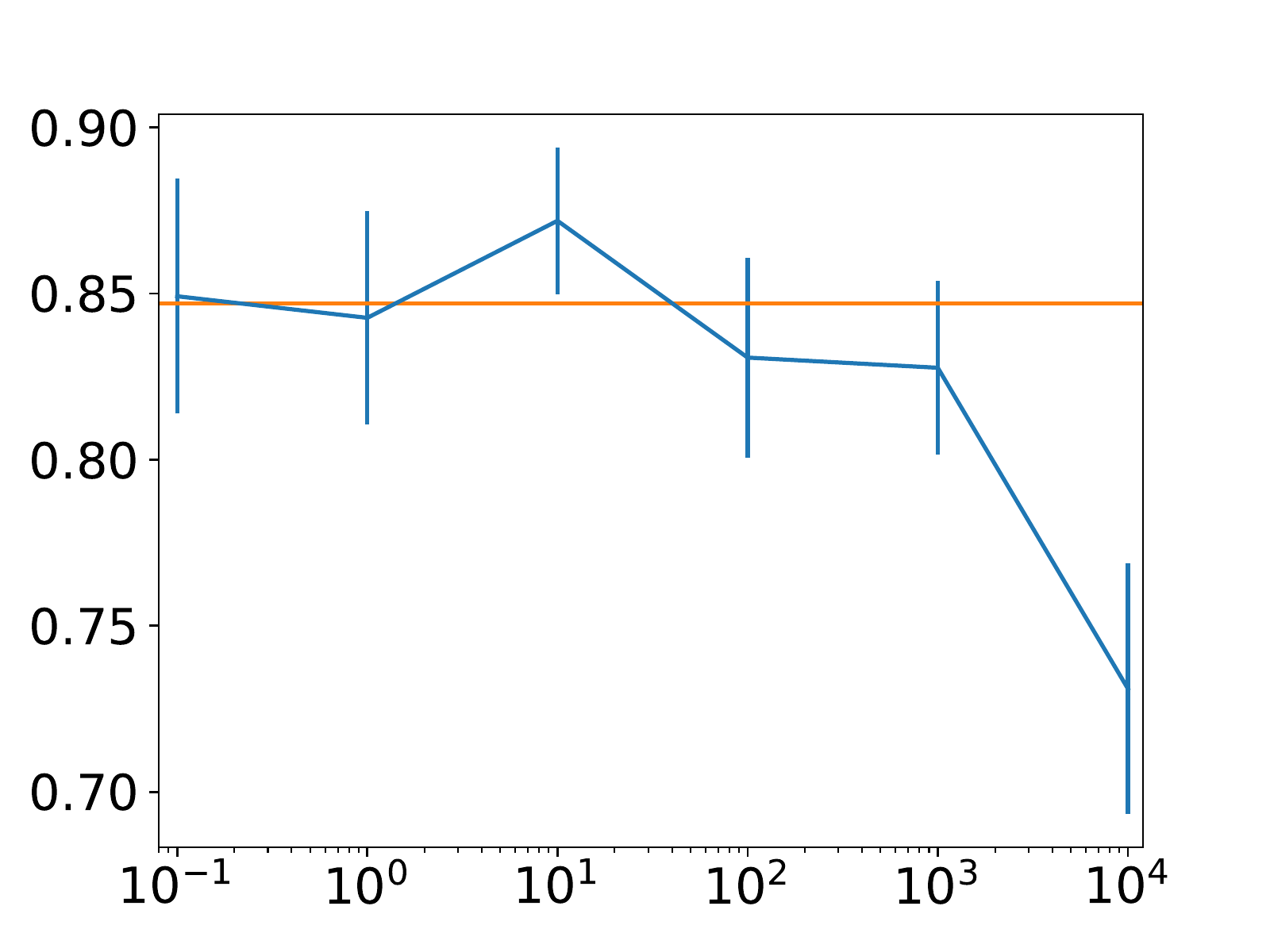}&
\includegraphics[width=8.3em]{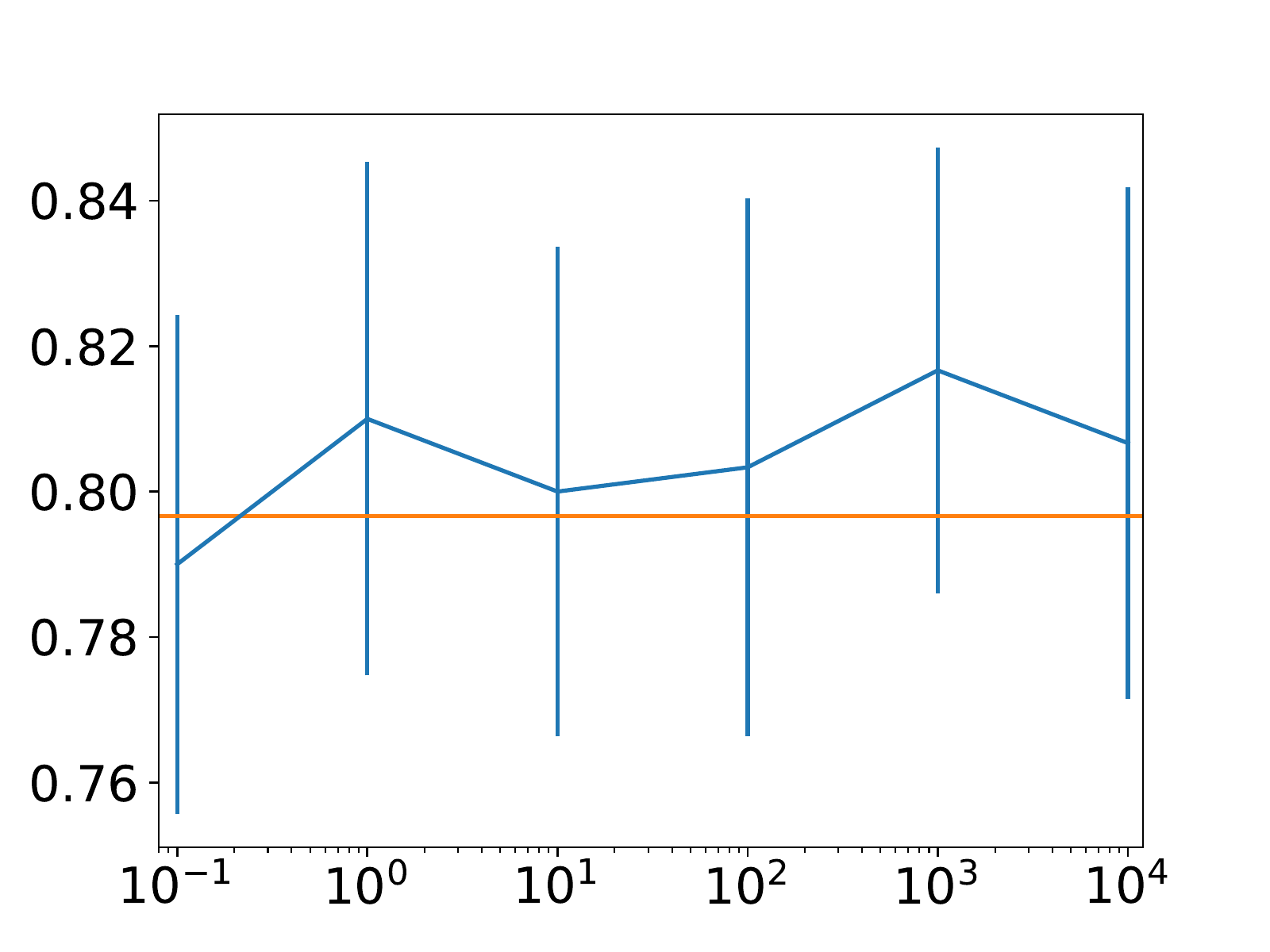} &
\includegraphics[width=8.3em]{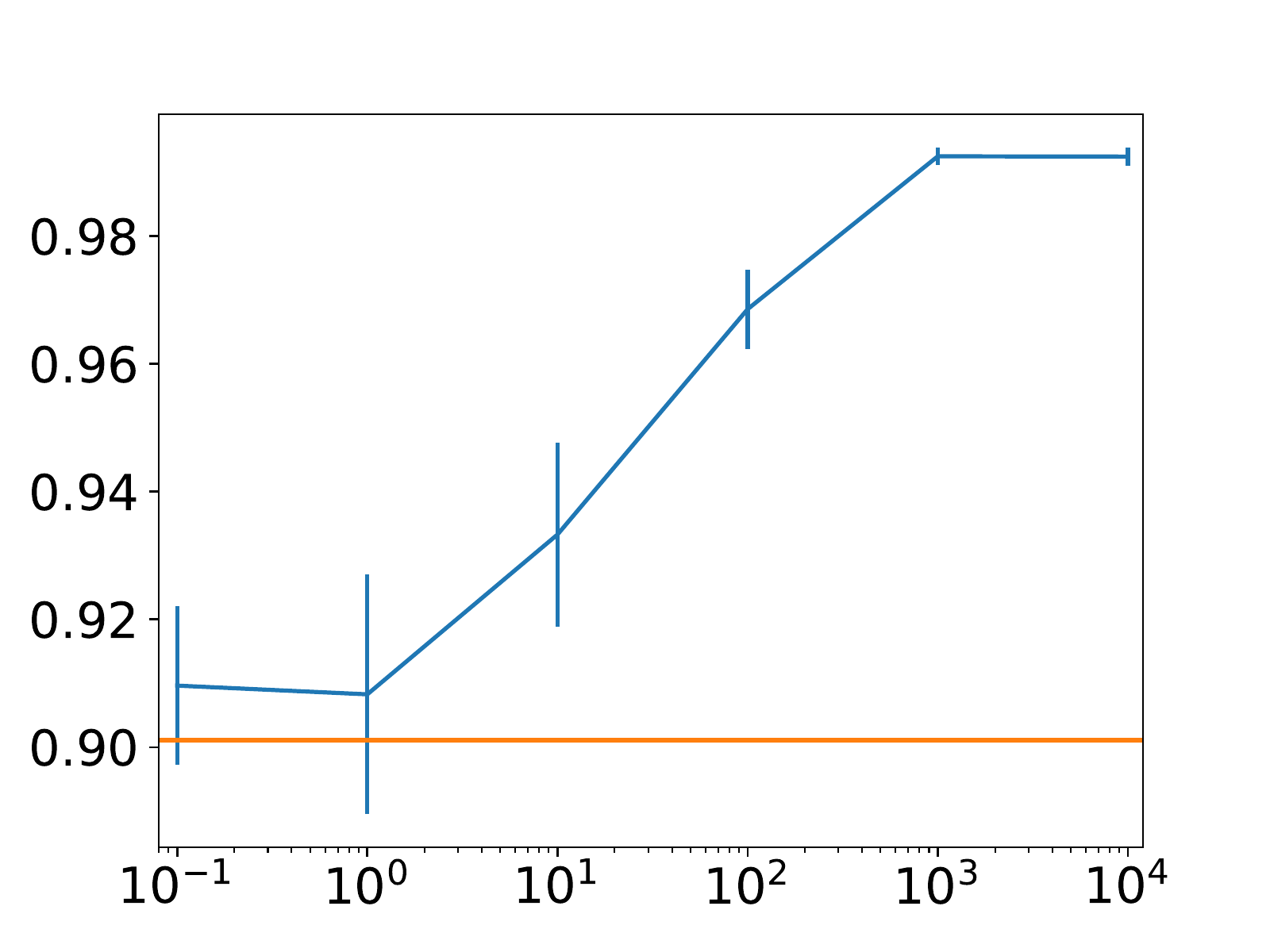} &
\includegraphics[width=8.3em]{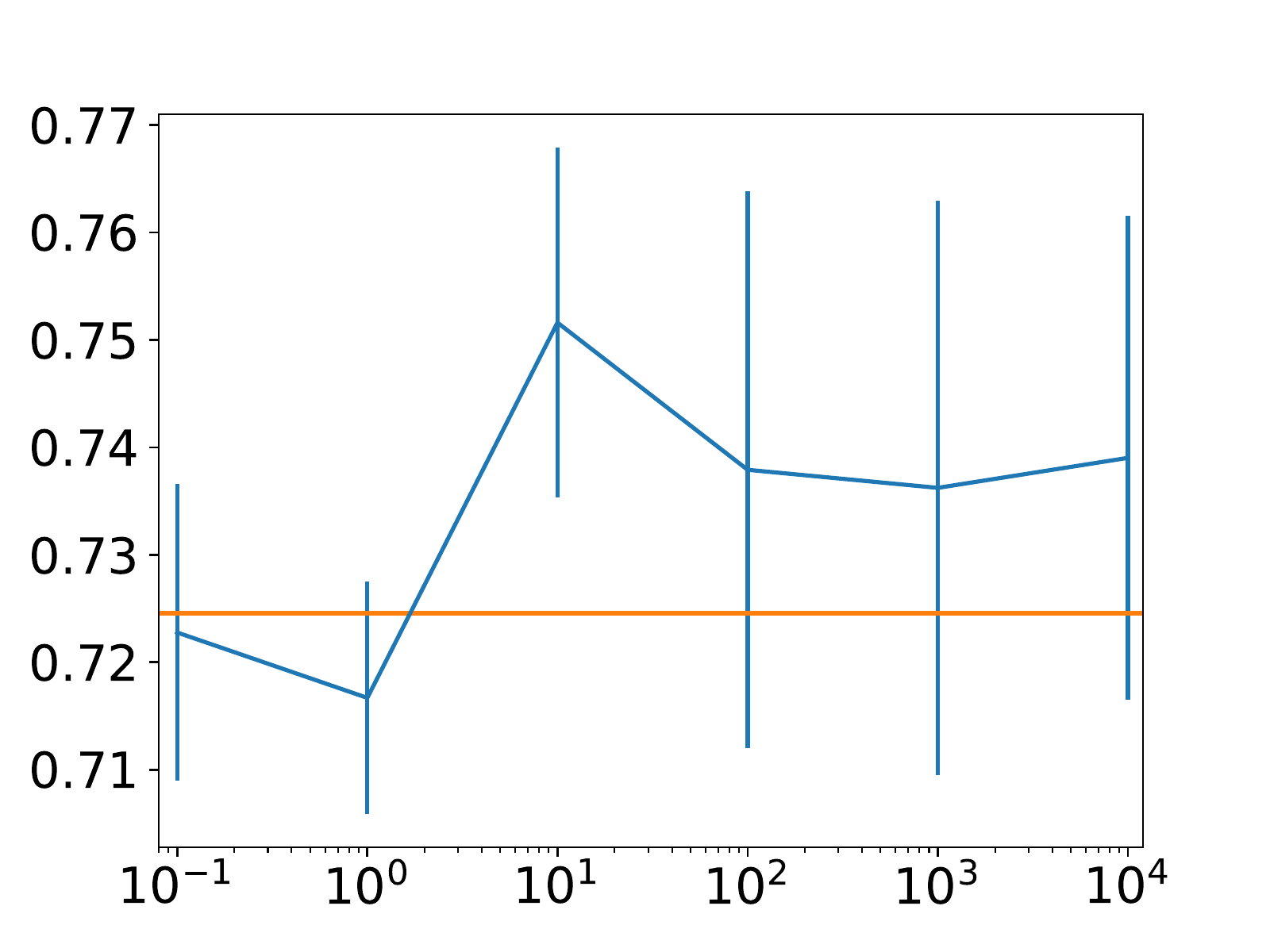} &
\includegraphics[width=8.3em]{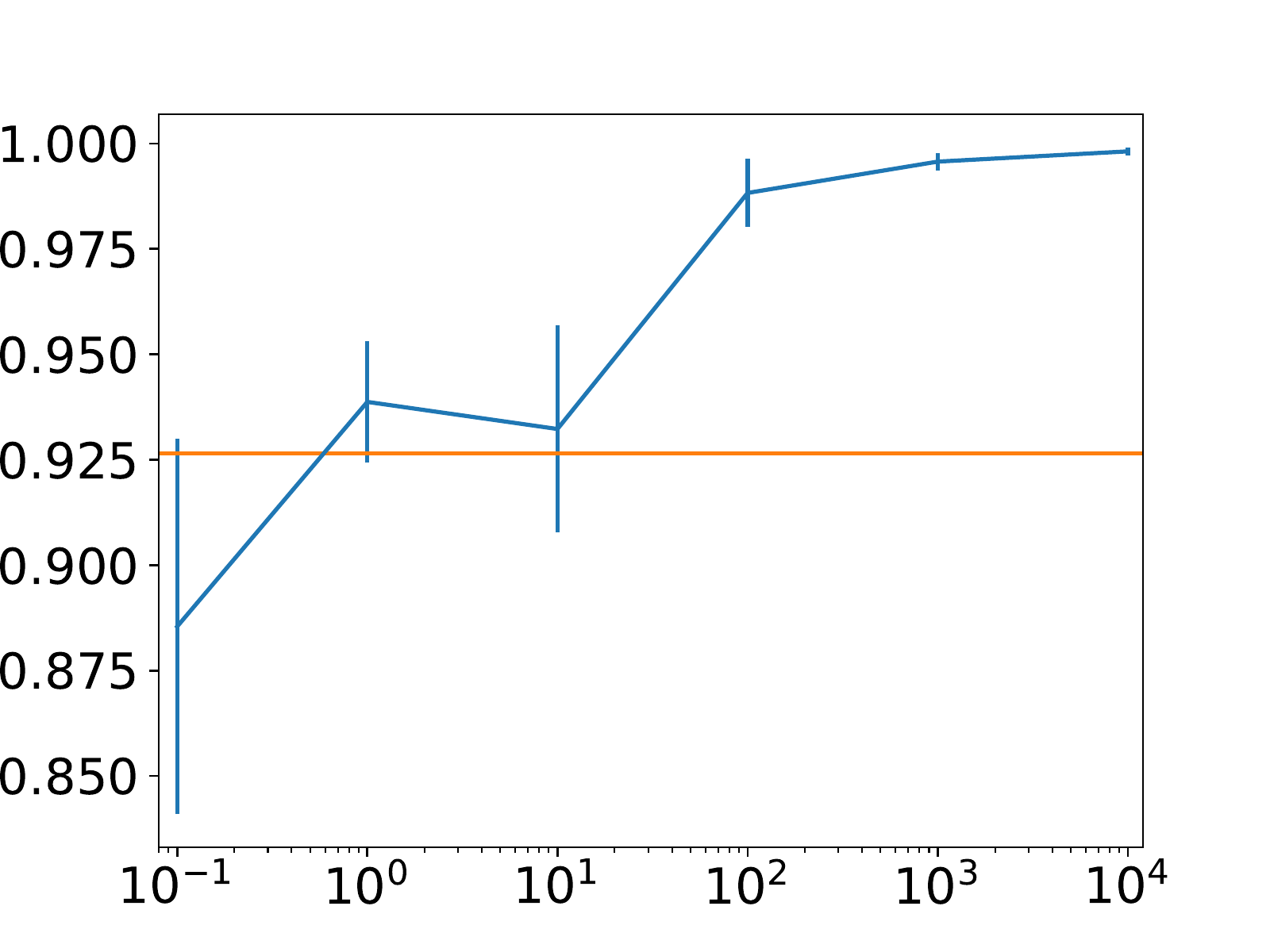} \\
KDDCup99&
PageBlocks&
Parkinson&
PenDigits&
Pima&
Shuttle\\
\includegraphics[width=8.3em]{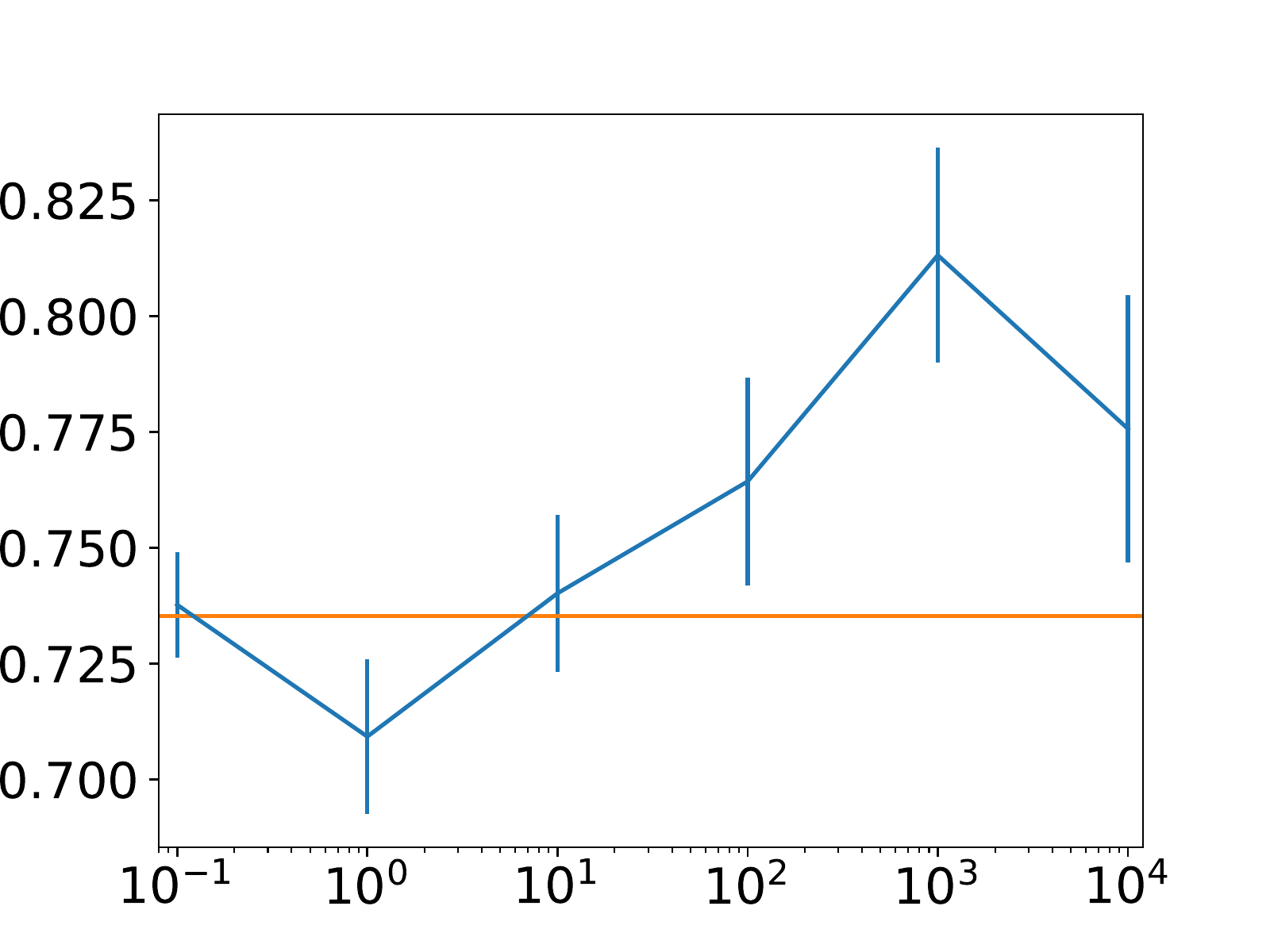} &
\includegraphics[width=8.3em]{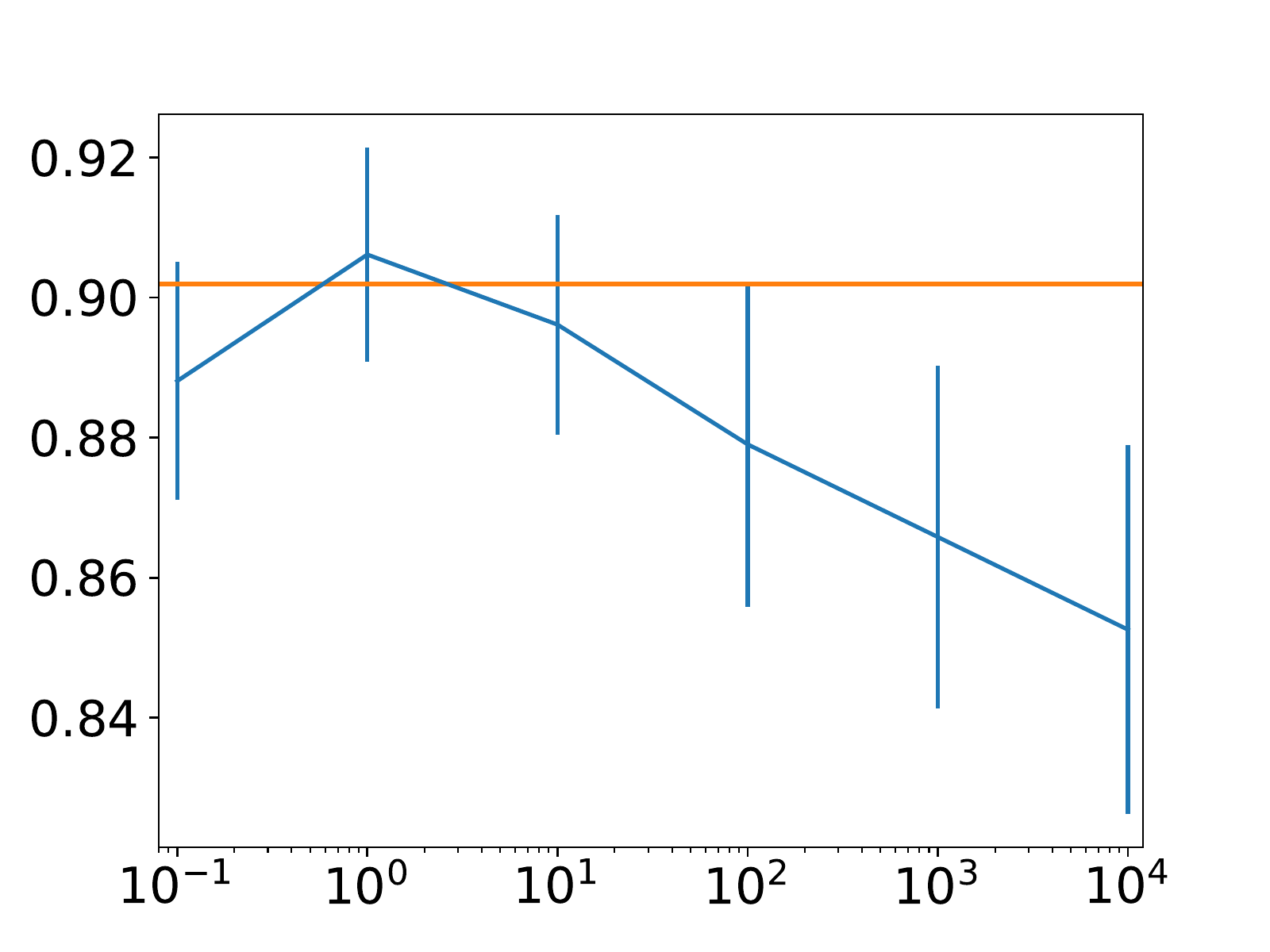} &
\includegraphics[width=8.3em]{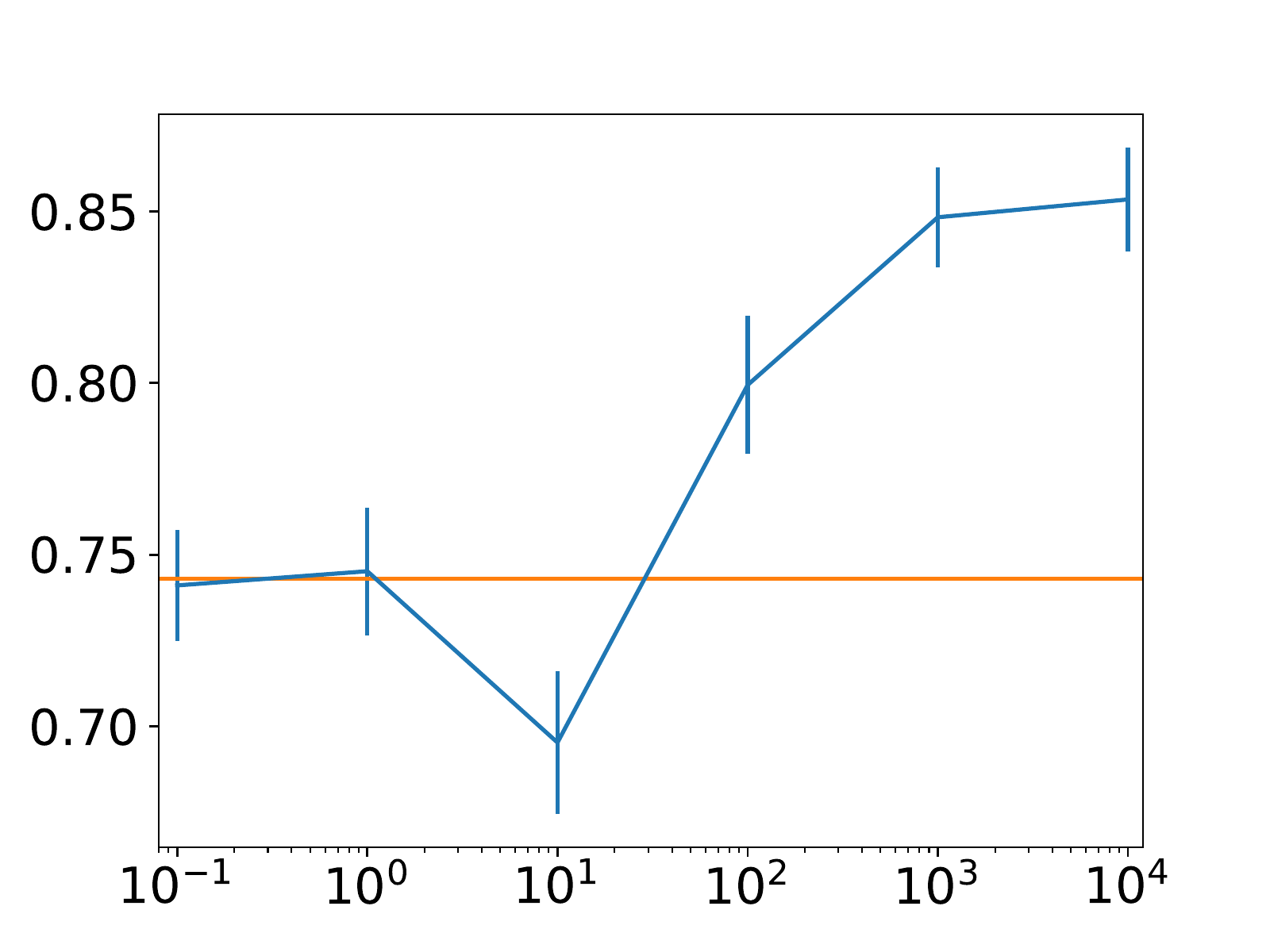} &
		\includegraphics[width=8.3em]{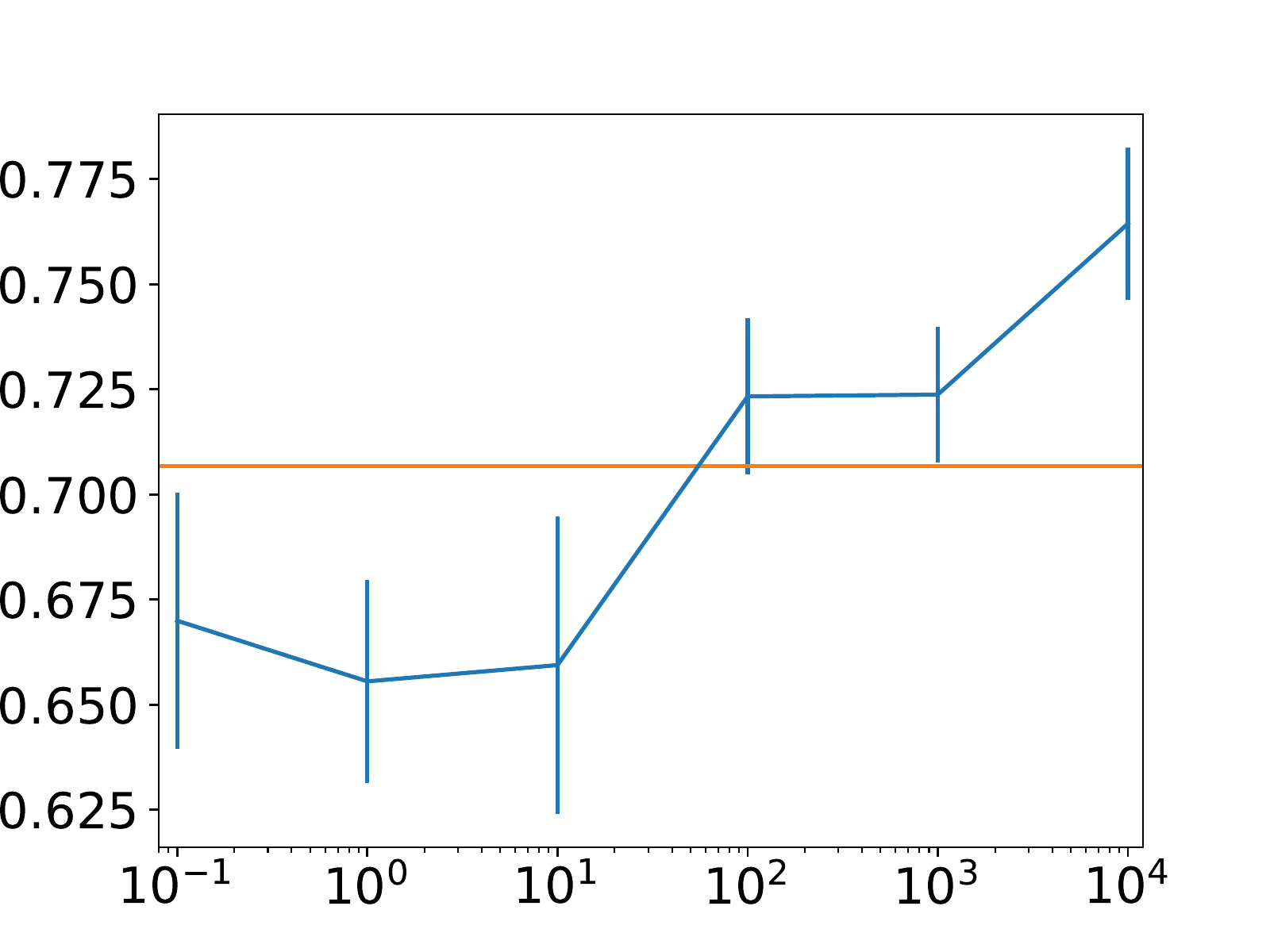} &&
			\\
SpamBase&
Stamps&
Waveform&
Wilt&
   \end{tabular}}
    \caption{AUCs by the proposed method with different hyperparameters $\lambda$. The x-axis is $\lambda$ and the y-axis is the AUC. The errorbar shows the standard error. The horizontal line is the AUC with $\lambda=0$.}
    \label{fig:lambda_auc}
   \end{figure*}

Table~\ref{tab:auc3} shows the AUC results.
The proposed method achieved the highest average AUC among the ten methods.
The AUC by the MADE was high
compared with the other unsupervised methods, which
indicates that the neural autoregressive density estimators would be useful
for unsupervised anomaly detection.
With some datasets,
the AUC by the proposed method was statistically higher than that by the MADE,
e.g. Annthyroid, PenDigits and Wilt.
There were no datasets where the AUC by the MADE was statistically higher than that by the proposed method.
This result indicates that the regularization term in the proposed method works well
for improving anomaly detection performance with a few labeled anomalous instances.
The SVM achieved the high AUC among supervised binary classifier based methods.
However, the performance of the SVM with some datasets was
very low, e.g. Arrhythmia and Pima.
On the other hand,
the proposed method achieved relatively high AUC on all of the datasets.
This would be because the proposed method incorporates the characteristics of unsupervised methods
by the likelihood term in the objective function as well as
the characteristics of supervised methods by the regularization term.
Table~\ref{tab:auc15} shows the AUC results with (a) one and (b) five training anomalous instances
by the supervised methods.
The proposed method also achieved the highest average AUC with these settings.
The average AUC by the proposed method
increased as the number of training anomalous instances increased,
where the AUCs were 0.807, 0.821, 0.839, 0.859 when zero, one, three, five anomalous instances were used for training.
The average computational time for training the proposed method was
2.78, 0.53, 0.24, 0.03, 9.50, 0.05, 62.39, 1.36, 0.02, 4.12, 0.05, 0.09, 2.06, 0.03, 0.99, 0.82
seconds on
Annthyroid, Arrhythmia, Cardiotocography, HeartDisease, InternetAds, Ionosphere, KDDCup99, PageBlocks, Parkinson, PenDigits, Pima, Shuttle, SpamBase, Stamps, Waveform, Wilt dataset, respectively,
using a computer with a Xeon Gold 6130 CPU 2.10GHz.

Figure~\ref{fig:lambda_auc} shows the AUCs on the test data
by the proposed method with different hyperparameters $\lambda$
trained on datasets with three anomalous instances. 
The best hyperparameters were different across datasets.
For example,
the high $\lambda$ was better with the PenDigits and Shuttle datasets,
the low $\lambda$ was better with the Annthyroid and Stamps datasets, and
the intermediate $\lambda$ was better with the Cardiotocography and Ionosphere datasets.
This result indicates that the AUC maximization without the likelihood maximization,
which corresponds to the proposed method with the high $\lambda$,
is not effective in some datasets.
The proposed method achieved the high performance with various datasets
by automatically adapting the $\lambda$ using the validation data
to control the balance of the likelihood maximization and the AUC maximization.
   
\section{Conclusion}
\label{sec:conclusion}

We have proposed a supervised anomaly detection method
based on neural autoregressive models.
With the proposed method, the neural autoregressive model is trained so that the likelihood of normal instances is maximized
and the likelihood of anomalous instances is lower than that of normal instances.
The proposed method can detect anomalies
in the region where there are no normal instances
as well as in the region where anomalous instances are located closely.
We have experimentally confirmed the effectiveness of the proposed method using 16 datasets.
Although our results have been encouraging to date, our
approach can be further improved in a number of ways.
First, we would like to extend our framework for
semi-supervised setting~\cite{blanchard2010semi},
where unlabeled instances as well as labeled anomalous and normal instances are given.
Second, we plan to incorporate
other neural density estimators including VAE to our framework.

\bibliographystyle{aaai}
\bibliography{aaai2019_anomaly}

\end{document}